# DESIGN OF A 3 AXIS PARALLEL MACHINE TOOL FOR HIGH SPEED MACHINING : THE ORTHOGLIDE

**Félix Majou**

Institut de Recherche en Communications et Cybernétique de Nantes
(IRCCyN : U.M.R. 6597, Ecole Centrale de Nantes, Ecole des Mines de Nantes, Université de Nantes)
1 rue de la Noë, BP 92101, 44321 Nantes Cedex 03, France.
Tel : 02 40 37 69 52, Fax : 02 40 37 69 30
Felix.Majou@irccyn.ec-nantes.fr

**Philippe Wenger**

Institut de Recherche en Communications et Cybernétique de Nantes,
1 rue de la Noë, BP 92101, 44321 Nantes Cedex 03, France.
Tel : 02 40 37 69 47, Fax : 02 40 37 69 30
Philippe.Wenger@irccyn.ec-nantes.fr

**Damien Chablat**

Institut de Recherche en Communications et Cybernétique de Nantes,
1 rue de la Noë, BP 92101, 44321 Nantes Cedex 03, France.
Tel : 02 40 37 69 54, Fax : 02 40 37 69 30
Damien.Chablat@irccyn.ec-nantes.fr

**Abstract :**

*The Orthoglide project aims at designing a new 3-axis machine tool for High Speed Machining. Basis kinematics is a 3 degree-of-freedom translational parallel mechanism. This basis was submitted to isotropic and manipulability constraints that allowed the optmization of its kinematic architecture and legs architecture. Thus, several leg morphologies are convenient for the chosen mechanism. We explain the process that led us to the choice we made for the Orthoglide. A static study is presented to show how singular configurations of the legs can cause stiffness problems.*

# 1 Introduction

## 1.1 Serial machine tools problems

Most industrial machine tools have a serial kinematic architecture, which means that each axis has to carry the following one, including its actuators and joints (Fig. 1). High Speed Machining (HSM) highlights some drawbacks of such architectures: heavy moving parts require from the machine structure high stiffness to limit bending problems that lower the machine accuracy, and limit the dynamic performances of the feed axes.

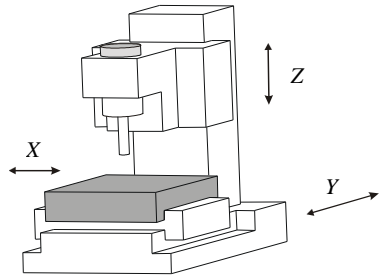

Fig. 1 : *A serial 3-axis Machine Tool*

## 1.2 Parallel Kinematic Machines (PKMs) are alternative machine tool designs for High Speed Machining

In a PKM, the tool is connected to the base through several kinematic chains or legs that are mounted in parallel. Figure 2 shows a 3-degree-of-freedom (3-DOF) planar mechanism ( 2 translations, 1 rotation) mounted on 3 RPR legs (Revolute, Prismatic and Revolute joints).

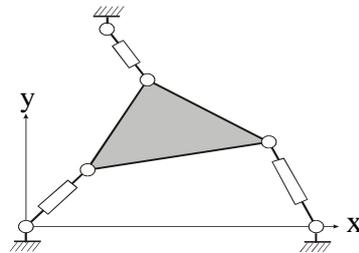

Fig. 2 : *A 3-RPR parallel mechanism*

PKMs attract more and more researchers and companies, because they are claimed to offer several advantages over their serial counterparts, like high structural rigidity and high dynamic capacities. Indeed, the parallel kinematic arrangement of the links provides higher stiffness and lower moving masses that reduce inertia effects. Thus, PKMs have better dynamic performances, which is interesting for HSM. Most existing PKMs can be classified into two main families: (i) PKMs with fixed foot points and variable strut lengths and (ii) PKM with fixed length struts and moveable foot points.

The first family is mostly composed of the so-called hexapod machines which, in fact, feature a Gough-Stewart platform architecture. Many prototypes and commercial hexapod PKMs already exist like the VARIAX-Hexacenter (Gidding&Lewis), the CMW300 (Compagnie Mécanique des Vosges), the TORNADO 2000 (Hexel), the MIKROMAT 6X (Mikromat/IWU), the hexapod OKUMA (Okuma), the hexapod G500 (GEODETIC). In this first family, we find also hybrid architectures : the TRICEPT 845 from Neos Robotics which is a 2-axis wrist serially mounted with a 3-DOF parallel structure, and a 3-DOF hybrid high speed machine tool designed by LARAMA and PCI [1].

The second family (ii) of PKM has been more recently investigated. In this category we find the HEXAGLIDE (ETH Zürich) which features six parallel (also in the geometrical sense) and coplanar linear joints. The HexaM (Toyota) is another example with non coplanar linear joints. A 3-axis translational version of the hexaglide is the TRIGLIDE (Mikron), which has three coplanar and parallel linear joints. Another 3-axis translational PKM is proposed by the ISW Uni Stuttgart with the LINAPOD. This PKM has with three vertical (non coplanar) linear joints. The URANE SX (Renault Automation Comau) and the QUICKSTEP (Krause & Mauser) are 3-axis PKM with three non coplanar horizontal linear joints. A hybrid parallel/serial PKM with three parallel inclined linear joints and a two-axis wrist is the GEORGE V (IFW Uni Hanover). H4, a family of 4-DOF parallel robots was presented in [2].

Most PKMs suffer from the presence of singular configurations in their workspace that limit the machine performances.

### 1.3 Singular configurations

The singular configurations (also called singularities) of a PKM may appear inside the workspace or at its boundaries. There are two main types of singularities [3]. A configuration where a finite tool velocity requires infinite joint rates is called a serial singularity or a type 1 singularity. A configuration where the tool cannot resist any effort and in turn, becomes uncontrollable, is called a parallel singularity or type 2 singularity. Parallel singularities are particularly undesirable because they cause the following problems:

- a high increase of forces in joints and links, that may damage the structure,
- a decrease of the mechanism stiffness that can lead to uncontrolled motions of the tool though actuated joints are locked.

Figures 3a and 3b show the singularities for a "biglide" mechanism, which is a 2-PRR mechanism with prismatic actuated joints. Its legs are made of fixed length struts with gliding node points. In Fig. 3a, we have a serial singularity. The velocity amplification factor along the vertical direction is null and the force amplification factor is infinite.

Figure 3b shows a parallel singularity. The velocity amplification factor is infinite along the vertical direction and the force amplification factor is close to zero. Note that a high velocity amplification factor is not necessarily desirable because the actuator encoder resolution is amplified and thus the accuracy is lower.

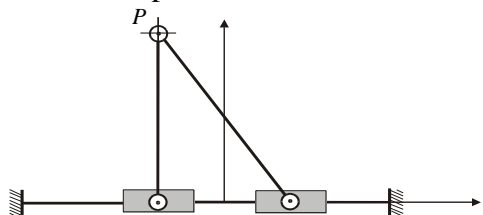
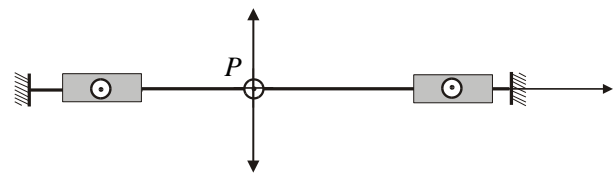

Fig. 3a : *A serial singularity*  Fig. 3b : *A parallel singularity*

### 1.4 The Orthoglide project

The Orthoglide project aims at building a prototype of a parallel kinematic machine tool for HSM with a kinematic behaviour similar to the one of a classical serial 3-axis machine. First, we present some research works on the structural synthesis of 3-DOF translational parallel mechanisms that helped our choice of the Orthoglide architecture. Then we show how isotropic and manipulability constraints led to modifications of the basis architecture as well as of the legs architecture. The study of forces inside the legs concludes this study.

## 2 Choice of a suitable kinematic architecture

### 2.1 Design of a 3-DOF translational parallel mechanism

Many studies have been conducted on the design of parallel mechanisms. Hervé proposed in [4] a tool for the synthesis of parallel robots based on the mathematical group theory. In [5], this tool was applied to the design of a 3-DOF translational parallel manipulator called Y-STAR (Fig. 4). The author in [6] explored tools for the design and optimization of parallel mechanisms with constrained DOFs. Recently, Kong proposed in [7] the generation of translational parallel manipulators based on screw theory [8].

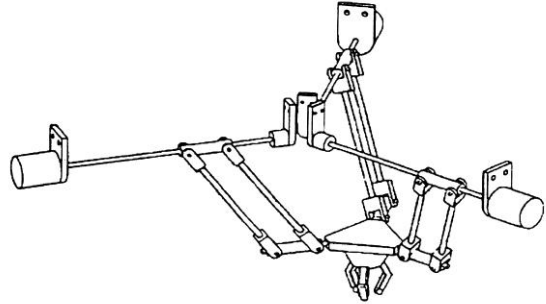

Fig. 4 : *The Y-Star manipulator*

### 2.2 A convenient joints architecture

A convenient PKM architecture for HSM has to respect some technological constraints :
- Only 1-DOF in kinematic links, for a simple design and a low cost.
- Actuators fixed on the frame, to reduce to the maximum inertia effects.
- Actuated prismatic joints to enable the use of linear motors.
- Similar legs, for a low cost.

The Y-Star robot, with helical actuated joints replaced by prismatic actuated joints followed by passive revolute joints is a convenient choice regarded to these constraints. We chose it as the basis mechanism of the Orthoglide project.

The structural synthesis is now achieved : an ordered set of joints is available for each leg. We now have to adjust it regarded to isotropic and manipulability constraints.

## 3. Optimization of the leg architecture

### 3.1 Conditioning index and manipulability

For a serial mechanism, the velocity and force transmission ratios are constant in the workspace. For a parallel mechanism, in contrast, these ratios may vary significantly in the workspace because the displacement of the tool is not linearly related to the displacement of the actuators. In some parts of the workspace, the maximal velocities and forces measured at the tool may differ significantly from the maximal velocities and forces that the actuators can produce. This is particularly true in the vicinity of singularities. At a singularity, the velocity, accuracy and force ratios reach extreme values.

Let $\dot{\rho}$ be referred to as the vector of actuated joint rates and $\dot{p}$ as the velocity vector of point P. $\dot{\rho}$ and $\dot{p}$ are related through the Jacobian matrix **J** as follows :

$$\dot{p} = \mathrm{J}\, \dot{\rho}$$

**J** also relates the static tool efforts to the actuator efforts. For parallel manipulators, it is more convenient to study the conditioning of the Jacobian matrix that is related to the inverse transformation, $\mathbf{J}^{-1}$. To evaluate the ability of a parallel mechanism to transmit forces or velocities from the actuators to the tool, two complementary kinetostatic performance criteria can be used : the conditioning of the inverse Jacobian matrix $\mathbf{J}^{-1}$, called conditioning index, and the manipulability ellipsoid associated with $\mathbf{J}^{-1}$ [9].

The conditioning index is defined as the ratio of the highest to the smallest eigenvalue of $\mathbf{J}^{-1}$. The conditioning index varies from 1 to infinity. At a singularity, the index is infinity. It is 1 at another special configuration called isotropic configuration. At this configuration, the tool velocity and stiffness are equal in all directions. The conditioning index measures the uniformity of the distribution of the velocities and efforts around one given configuration but it does not inform about the magnitude of the velocity amplification or effort factors.

The manipulability ellipsoid is defined by its principal axes as the eigenvectors of $(\mathbf{J}\,\mathbf{J}^T)^{-1}$ and by the lengths of the principal axes as the square roots of the eigenvalues of $(\mathbf{J}\,\mathbf{J}^T)^{-1}$. The eigenvalues are associated with the velocity (or force) amplification factors along the principal axes of the manipulability ellipsoid [9].

**3.2 Design constraints imposed to cope with this problem**

To design a translational PKM with kinematic behaviour close to the one of a serial 3-axis machine tool, we impose the following conditions : (i) there is one point in the workspace for which the velocity transmission ratio is the same in every direction, (ii) and its value is one at this configuration. In [10]  and [11]  the geometric conditions implied by these constraints are described in a more rigorous way.

Condition (i) means that there is an isotropic configuration in the workspace.

Condition (ii) means that for this configuration, the velocity amplification factors along the principal axes of the manipulability ellipsoid are equal to 1.

**3.3 Geometrical arrangement of the legs**

These two kinetostatic conditions lead to new geometric conditions on the Y-Star legs :
- (i) implies that for each leg (Figure 5), orientation between the axis of the linear joint $\mathbf{T_i}$ and the axis of parallelogram $\mathbf{W_i}$ must be the same for each leg i, and that all vectors $\mathbf{W_i}$ must be orthogonal to each other [10].

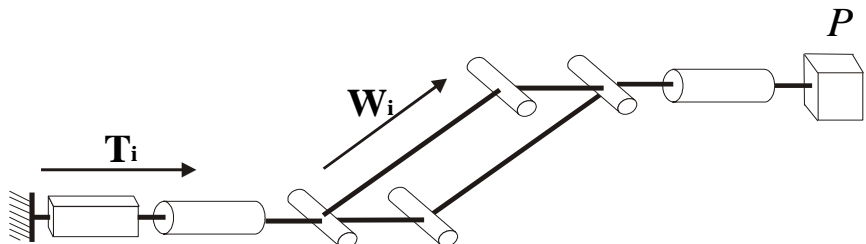

Fig. 5 : *Y-Star leg*

- (ii) implies that for each leg, $T_i$ and $W_i$ must be collinear [10]. Since, at the isotropic configuration, $W_i$ vectors are orthogonal, this implies that $T_i$ vectors are orthogonal, i.e. the linear joints are orthogonal (Figure 6).

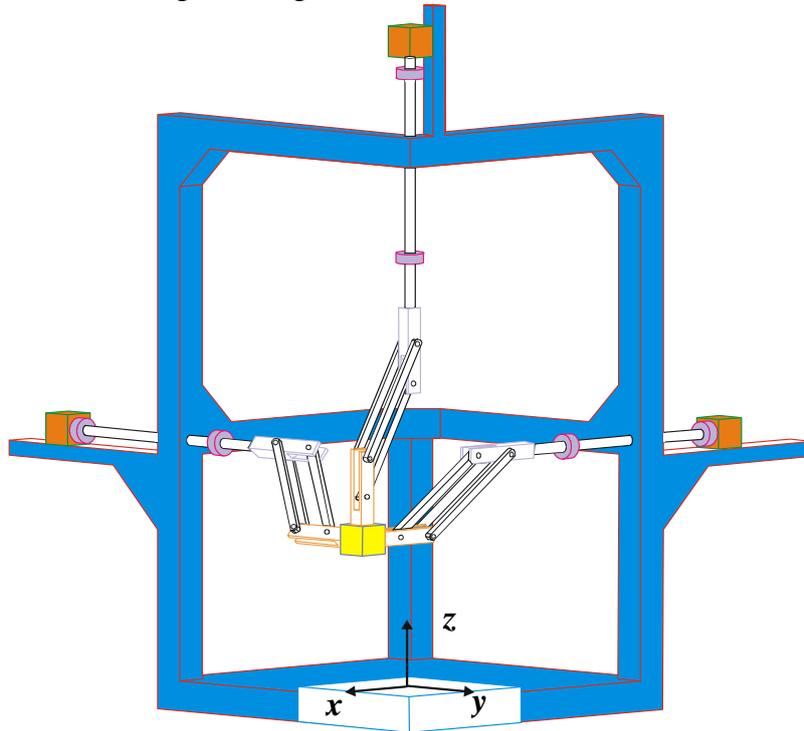

Fig. 6 : *New arrangement of Y-Star robot*

### 3.4 Rearrangement of legs architecture

The new geometrical arrangement of the machine legs leads to a singularity of the parallelogram (Figure 7) which becomes an antiparallelogram [12] : a passive rotation appears around an axis orthogonal to the parallelogram plane. A solution to this problem is to change the leg architecture by rearranging the joints, while keeping the same degree of freedom.

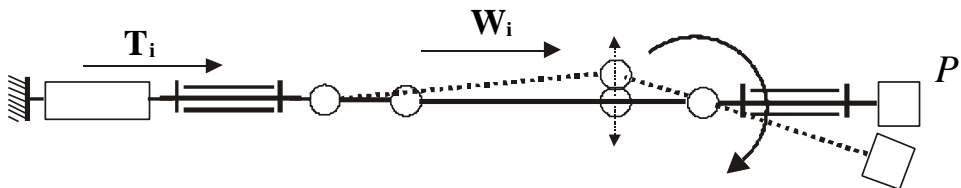

Fig. 7 : *Parallelogram singularity at isotropic configuration*

A second version of the leg architecture is then proposed (Figure 8a).

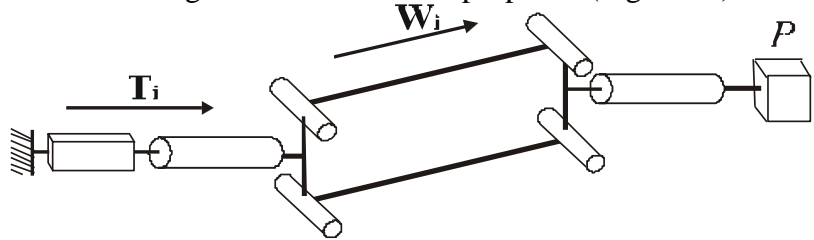

Fig. 8a : *Second version of the leg architecture*

The parallelogram singularity at the isotropic configuration is avoided but another problem arises : a special singularity of the leg at the isotropic configuration. There is a passive rotation around the $T_i$ axis. It appears that this particular singularity is not detected by the method described in [3]. [13] proposes a way to find it, and from [14] this type of singularity is a (RPM, IO, II) singularity. RPM means that a Redundant Passive Motion is possible. This motion is the rotation of the parallelogram around the $T_i$ axis (Figure 8b). IO and II mean that in this configuration we have an Impossible Output ($\dot{p} = 0$) and an Impossible Input ($\dot{\rho} = 0$), respectively.

At the isotropic configuration, each leg can passively transmit a force whose axis is orthogonal to the parallelogram plane which means that no translation of point *P* along this axis is possible (IO and II, see figure 8b).

Furthermore, to have a pure translational 3-DOF mechanism (the tool cannot rotate), the parallelograms must be orthogonal to one another at the isotropic configuration. Thus, the mechanism gets locked at this configuration because no translation nor rotation of the tool is possible.

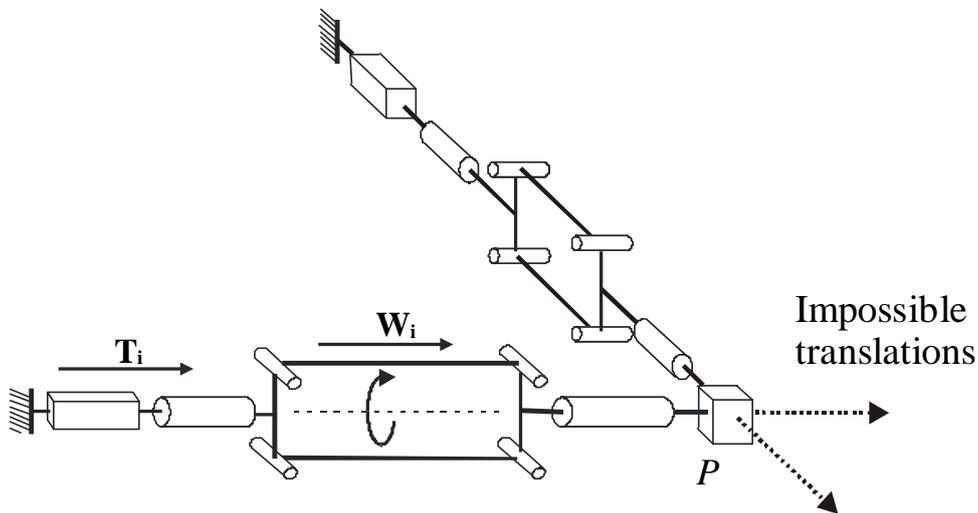

Fig. 8b : *Leg singularity at isotropic configuration*

The last version of Orthoglide legs (Figure 9) avoids all previous problems mentioned at the isotropic configuration : no parallelogram singularity, and no leg singularity.

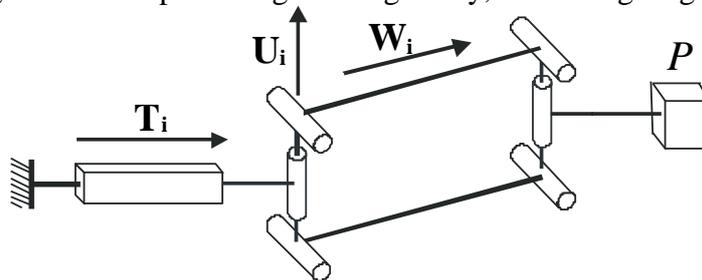

Fig. 9 : *Current version of Orthoglide legs*

### 3.5 Leg orientation for a pure translational mechanism

Now that a correct leg architecture has been defined, we have to choose the parallelograms orientation in the machine frame. The screw theory [8] allows a geometric explanation of the conditions on legs orientation that lead to a pure translational motion of the tool : the wrench system (forces or torques that can be passively transmitted by a kinematic chain) of each leg is composed of two torques (Figure 9) : the first torque axis is $T_i$, and the second torque axis is perpendicular to the plane ($T_i$, $U_i$). The tool can rotate when its wrench system (which is the union of all legs wrench systems) does not contain any torque. Consequently, the three legs wrench systems must include all torques, i.e., the three parallelograms planes must be orthogonal to one another. The result is shown on Figure 10. Note that the constant orientation of parallelograms in the whole workspace makes the Orthoglide free from constraint singularities[1].

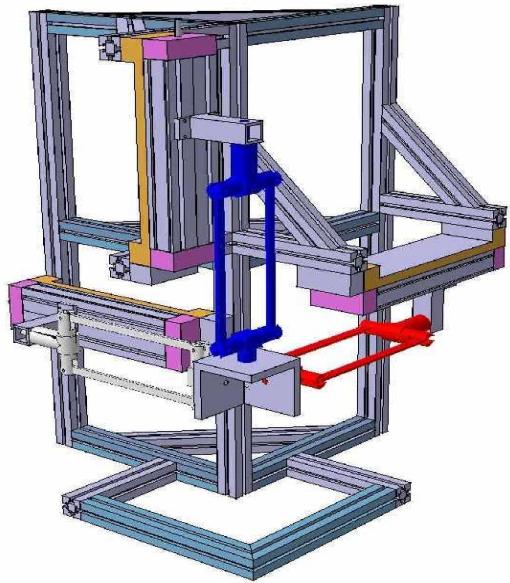

Fig. 10 : *Orthoglide legs orientation*

### 4. Static analysis of legs parallelograms

At a parallelogram singularity (see figure 11 and 12), a passive rotation of bar 3 around axis ($B_i$, $S_i$) appears, though input and output motions are "locked". At this singularity, the tension / compression forces in bars 1 and 2 tend to infinite. Physically, bar 3 can not be statically balanced, therefore a motion is possible.

---

[1] Constraint singularities were discovered recently [15] [16]. They may arise for parallel mechanisms with less than 6 DOFs. At a constraint singularity, the moving platform gains a new DOF.

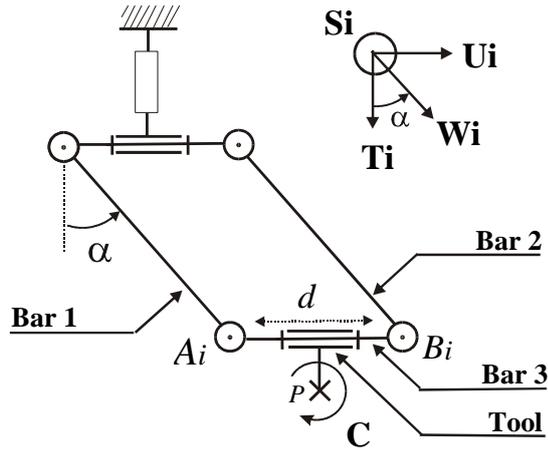 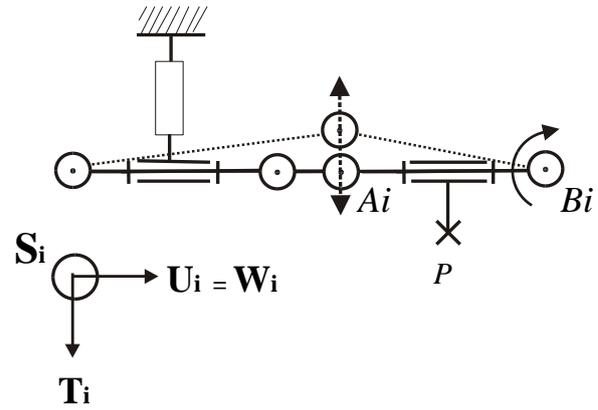

Fig. 11 : *Static model of the parallelogram*   Fig. 12 : *Parallelogram singularity*

The parallelogram on figure 11 is in the plane ($T_i$, $U_i$) because the orientation around $U_i$ has no influence on the tension / compression forces in bars 1 and 2. Furthermore, we assume that only one torque **C** is exerted on the tool at point *P* : **C** = C.$S_i$

The parallelogram is statically balanced when $\alpha \neq 90°$ and only tension / compression forces are generated in bars 1 and 2. The force exerted by bar 1 on bar 3 is : $F_b$ = $F_b$.$W_i$

The force exerted by bar 2 on bar 3 is the opposite. These forces are fully transmitted to the tool by the revolute joint around $U_i$.
The static equation of the torque exerted on the tool at point *P* can be written as follows :

$$2 \times [(F_b \times \cos\alpha) \times (d/2)] = C$$
$$\Rightarrow F_b = (C/d) \times 1/\cos\alpha$$

When the parallelogram approaches its singularity, we have : $\alpha \to 90°$ and $F_b \to \infty$. The tension / compression forces in bars 1 and 2 at joints limits have to be checked. In [11] the maximum value of $\alpha$ was calculated as : $\alpha_{max}$ = 14°.

Our design gave us : *d* = 100 mm. An approximated value of the torque C exerted on the tool regarded to machining conditions expected from our prototype is : C ≈ 10 Nm. Thus, the tension / compression force in bars 1 and 2 is : $F_b$ = 103 N. The section of bars 1 and 2 is : S = 144 mm². The maximum tensile stress in bars 1 and 2 is then : $\sigma_{max}$ = 0,7 Mpa. This result is far less than the legs material maximum tensile strength (aluminium).

## 5. Conclusions

This paper describes the design of a new 3-axis machine tool based on a translational parallel mechanism : the Orthoglide. We chose a convenient architecture regarded to technological constraints. To have a kinematic behaviour close to the one of a serial 3-axis machine tool, we optimized the kinematic architecture to fit with isotropic and manipulability constraints. Several leg architectures were convenient. We chose the correct architecture by eliminating all legs singularities, particularly those that could not be detected by the classical velocities equations. Tensile strength in the parallelograms bars was evaluated for the most penalizing configurations.